\documentclass[letterpaper, 10 pt, conference]{ieeeconf}  

\IEEEoverridecommandlockouts                              

\usepackage{amsmath} 
\usepackage{amssymb}  
\usepackage{graphicx}
\usepackage{multirow}
\usepackage{multicol}
\usepackage{xcolor}
\usepackage{comment}
\title{\LARGE \bf
Probabilistic Multi-modal Trajectory Prediction with Lane Attention for Autonomous Vehicles
}

\author{Chenxu Luo$^{1,2}$,Lin Sun$^{2}$, Dariush Dabiri$^{2}$ and Alan Yuille $^{1}$
\thanks{*This work was done when Chenxu Luo was an intern at Samsung}
\thanks{$^{1}$Chenxu Luo and Alan Yuille are with Department of Computer Science, Johns Hopkins University, Baltimore, MD 21218.
        {\tt\small chenxuluo@jhu.edu, ayuille1@jhu.edu}}%
\thanks{$^{2}$Lin Sun and Dariush Dabiri are with Samsung Strategy and Innovation Center, Samsung, Inc., 3655 North First Street, San Jose, CA, 95134.
        {\tt\small lin1.sun@samsung.com, d.dabiri@samsung.com}}%
}

\begin{document}

\maketitle
\thispagestyle{empty}
\pagestyle{empty}

\begin{abstract}
Trajectory prediction is crucial for autonomous vehicles. The planning system not only needs to know the current state of the surrounding objects but also their possible states in the future. As for vehicles, their trajectories are significantly influenced by the lane geometry and how to effectively use the lane information is of active interest. Most of the existing works use rasterized maps to explore road information, which does not distinguish different lanes. In this paper, we propose a novel instance-aware representation for lane representation. By integrating the lane features and trajectory features, a goal-oriented lane attention module is proposed to predict the future locations of the vehicle. We show that the proposed lane representation together with the lane attention module can be integrated into the widely used encoder-decoder framework to generate diverse predictions. Most importantly, each generated trajectory is associated with a probability to handle the uncertainty. Our method does not suffer from collapsing to one behavior modal and can cover diverse possibilities. Extensive experiments and ablation studies on the benchmark datasets corroborate the effectiveness of our proposed method. Notably, our proposed method ranks third place in the Argoverse motion forecasting competition at NeurIPS 2019~\footnote{https://ml4ad.github.io/}. 


\end{abstract}
\section{Introduction}
Empowering robotics to have the ability to imitate human intelligence to forecast future positions is essential to develop socially compliant robots or self-driving cars. For instance, according to the predicted trajectories of the vehicle in front of the ego vehicle, autonomous vehicles can answer the question ``do I need to yield” confidently and then make a proper motion planning to avoid the collision. This is a crucial feature for autonomous vehicles. Meanwhile, trajectory prediction is an extremely challenging task given the inherent uncertainty of the future and complex environment.

The behaviors of vehicles are mostly constrained by the road geometry. When driving, people usually pay attention to one or more lanes, which are their intentions or goals. They will either follow the current lane or drive towards an intended lane. Thus, modeling the lane information is crucial for vehicle trajectory prediction, especially for urban driving scenarios, where the number and the directions or the lanes can be varied over time. Previous works mainly~\cite{c6,c7} use rasterized maps as representation, which do not distinguish different lanes. In this paper, we first propose a novel instance-aware lane encoder to represent each lane individually. Each lane is represented by a set of ordered points sampled along the centerline, which is more memory-efficient than using rasterized map. The lane encoder extract a fixed-length vector for each lane representation. Based on that, we devise a novel goal-oriented lane attention module to predict the probability of a lane that the vehicle will enter onto in the next few seconds. This module is based on the dot-product attention which can deal with various number of lanes flexibly. It can also lead to semantic concepts, such as ``turn left" or ``change the lane". 

For the autonomous driving system, there might be several possible paths given the current observation of the vehicle and we can never be sure about the drivers' intention. Taking multiple plausible trajectories into consideration is essential for reliable motion planning. 
With the evolutionary progress of deep learning techniques, neural networks have already been applied to the trajectory prediction. Previous works \cite{c4, c5,mfp} mainly use GANs or VAEs to generate multiple hypotheses by sampling from random latent variables. However, two obvious drawbacks can be seen from this kind of method: (i) it is hard to determine the number of samples to cover all possible outcomes in practice. (ii) they treat all the predictions equally without assigning a reasonable probability for each one. Without a doubt, some trajectories are more likely to happen than others. 

In this paper, we show that our proposed lane attention model can be used to generate diverse and reasonable trajectories with probabilities. The lane intention provides reasonable priors for trajectory prediction. Based on the trajectory features, interaction features and the lane attention, we can generate the possible trajectory towards the ``goal''. Our method use the lane code as the latent variables for generating diverse trajectories, which is arguably more interpretable and can effectively avoid converging to the average trajectory. Compared with intention-based methods~\cite{socialstm,intentnet}, which use a fixed set of predefined classes, our lane attention module can be viewed as dynamic intention prediction to cover all the possibilities for the future. 

Our method unifies intention based and multi-modal trajectory prediction. We evaluate our method on a large real-world urban driving dataset Argoverse~\cite{argo} to show the effectiveness of our proposed method. 

We summarize our contributions as follows: 
\begin{itemize}
    \item We propose a novel instance-aware representation for lanes and a goal-oriented lane attention module for dynamic intention distribution prediction. 
    \item We provide a unified architecture for intention based and multi-modal trajectory prediction. 
    \item Extensive experiments and ablation studies on the benchmark dataset are conducted and verify the effectiveness of our proposed method. Notably, our proposed method ranks third place in the Argoverse motion forecasting challenge and we perform much better for the non-straight trajectories. 
\end{itemize}

\section{Related Works}
For prediction, Kalman Filter~\cite{kalman} is one of the widely used approaches for modeling uncertainty in prediction. With the remarkable achievements of deep learning, recently trajectory prediction algorithms also widely choose neural networks. Among all the choices, Long Short-Term Memory (LSTM) \cite{c8} and Gated Recurrent Unit (GRU) \cite{gru} are the most used networks to encode the temporal series \cite{lstm-encoder, socialstm, scene-lstm, sequence-lstm}. 

The future trajectories do not solely rely on the trajectory itself. They are greatly influenced by the interaction with other agents as well as with the static environment. There have been a lot of works for modeling interaction with other agents~\cite{socialstm,starnet,cslstm,nlpooling}, for either pedestrian or vehicle trajectory prediction. Some works also incorporate environment features into the models. Most of them apply a convolutional network to the image or the HD maps to extract environment cues~\cite{intentnet,c0,c6,c7,huang}. As for vehicles, the trajectories are greatly constrained by the surrounding lanes, especially in the complex urban environment. How to better model the interaction between the agents and the environment is still an open problem. 

Based on the number of output, trajectory prediction models can also be categorized into uni-modal and multi-modal predictions. 

Uni-modal models~\cite{c0, c1, c2,socialstm} only output the most likely trajectory, leaving most of the possible collision space unexplored which leads to an unreliable prediction. Moreover, they are prone to collapse to the average behavior modal which may not contribute to the valid prediction.

Generating multi-modal predictions~\cite{c4, c5, c6, c7} is essential, especially for autonomous vehicles. It can fully represent the vehicle behavior prediction space which can largely reduce the safety issues in motion planning. Some methods~\cite{c4,c5,c6} use VAEs or GANs to sample diverse predictions from latent random variables. One drawback of this kind of method is that it is not easy to get the probability associated with each prediction. The latent variables are less interpretable and it is also vulnerable to converge to one behavior modal as well.  

In order to handle these issues, 
~\cite{g1,g2,mdn} utilize guassian mixture model or mixture density network.
\cite{c7} directly predict six possible trajectories and their probability. However, without regularization, there is no guarantee to cover all the possibilities. 
Social-LSTM~\cite{socialstm} first predict the intention and then generate a trajectory for each intention. 
MultiPath~\cite{c6} leverages a fixed set of future state-sequence anchors that correspond to modes of trajectory distribution. In the paper, they applied k-means on the specific dataset to approximate the clusters and then uniformly sample trajectory space to obtain the anchors. Although effective, 
it is invariant to the agent’s current state or environment and needs a large number of anchors to cover all the outcomes.

\begin{figure*}[!h]
	\begin{center}
		\includegraphics[width=1.0\linewidth]{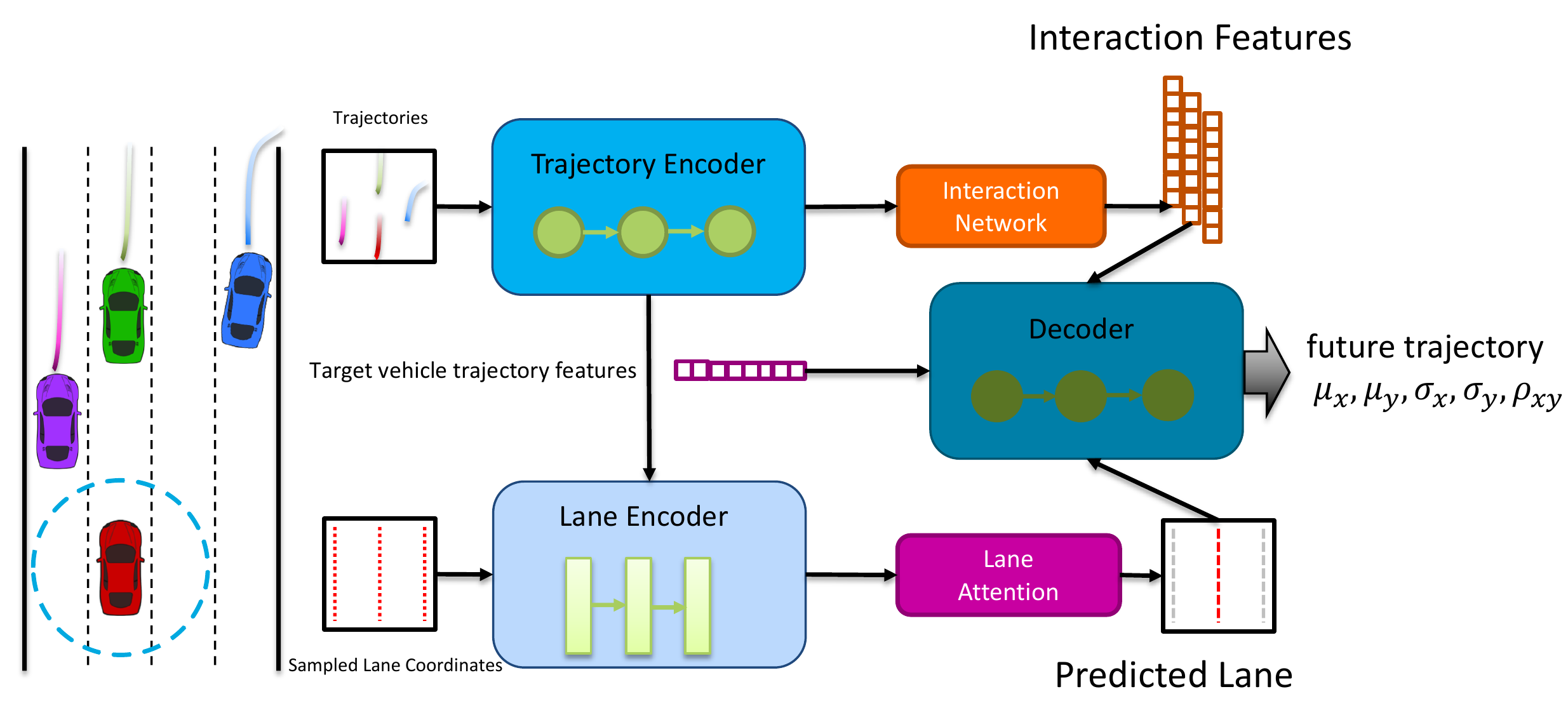}
	\end{center}
	\caption{An overview of our proposed method. The model consists of a trajectory encoder, a lane encoder, an interaction network, a lane attention module and a final trajectory decoder. Better to be viewed in color and zoom in. }
	\label{fig:overview}
\end{figure*}

\section{Proposed Algorithm}
In this section, we will describe our novel architecture which encodes the trajectories with lane attention for trajectory prediction. 
\subsection{Problem Formulation}
Given an observed vehicle trajectory histories within $t_o$ period $\mathcal{X}=(\mathbf{x}_{t_1},\mathbf{x}_{t_2},\cdots,\mathbf{x}_{t_o})$, vehicle velocity histories $\mathcal{V}=(\mathbf{v}_{t_1},\mathbf{v}_{t_2},\cdots,\mathbf{v}_{t_o})$, as well as corresponding histories of its surrounding agents and map information $\mathcal{M}$, the goal is to predict a $k$ sets of possible trajectories $\mathcal{Y}^k=(\mathbf{y}^k_{t_o+1},\cdots, \mathbf{y}^k_{t_h})$ in the future $t_h - t_o$ as well as the corresponding probability $p_k$.

\subsection{Architecture}
An overview architecture of our proposed model is shown in Figure \ref{fig:overview}. The system consists of five major modules, trajectory encoder, lane encoder, an attention module for lane selection, interaction network for intertwined agents and a trajectory decoder to generate future trajectories. The input will be the set of trajectories and sampled lane coordinates. The sets of trajectories will be fed into the trajectory encoder, extracting feature representations for each agent. The ego-vehicle trajectory features together with the sample lane coordinates will pass through a lane encoder to obtain lane attention. Finally, the interaction features and the predicted lane will pass through the encoder network to generate the future trajectories with probabilities associated with lanes.  

\subsection{Trajectory Encoding}
To encode the motion of the agent, similar to~\cite{socialstm}, we use the standard LSTM~\cite{c8}. We find that both the position and the velocity of the agent are important for prediction, therefore, in the architecture, our trajectory encoder consists of two LSTMs. One encodes position histories and the other one encodes velocity histories and then the two extracted features are concatenated as the agent motion feature. 
\begin{equation}
f_{{TR}_\text{i}}= g_1(\mathcal{X}^i) || g_2(\mathcal{V}^i)
\end{equation}
where $\mathcal{X}^i$ and $\mathcal{V}^i$ are the sequence of position and velocity for $i$-th agent, respectively. $g_1$ and $g_2$ are the LSTM modules for position and velocity. $f_{TR_{i}}$ is the output motion feature and $||$ is the concatenation. 

\subsection{Instance-Aware Map Encoding}
Previous works incorporate maps in trajectory prediction task by feeding rasterized maps into a convolutional network. The extracted map features will work together with the trajectory features for accurate prediction. However, it is not easy for the global map feature to distinguish different lanes explicitly. In this paper, we propose an instance-aware method to encode each lane using a shared network.

Specifically, we first search for all possible lanes within a certain distance surrounding the agent. Each lane is represented as a set of ordered points sampled from the center line. Figure~\ref{fig:lane} (a) illustrate an example of possible lanes around the vehicle (blue dot). 
The Lane Encoder is composed of two 1D convolutional layers and one multiple layer perceptron (MLP), as illustrated in Figure~\ref{fig:lane} (b). The final output is pooled to a fixed size feature vector $[f_{l_1}, f_{l_2} \cdots f_{l_n}]$ for each lane and $n$ is the number of lanes. 

Compared with the rasterized map, our method requires much less memory and our lane encoder is extremely lightweight.

\subsection{Goal Oriented Attention}
\begin{figure*}[!h]
	\begin{center}
		\includegraphics[width=0.9\linewidth, height=6cm]{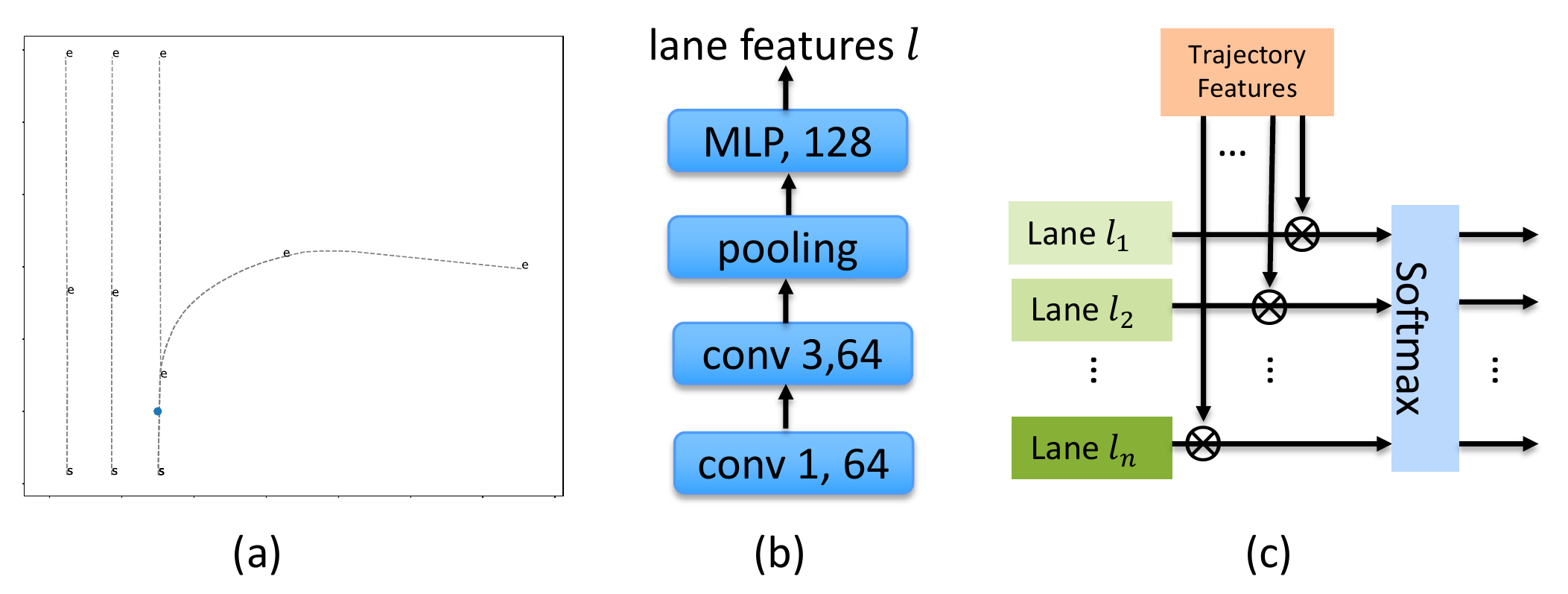}
	\end{center}
	\caption{(a) An example of selected lanes. The blue dot represents the last location of the target vehicle. ``s'' and ``e'' denotes the start and end of a road segment respectively. (b) The architecture of Lane Encoder. ``conv 1,64'' means 1D convolution with kernel size of 1 and 64 output channels. The final output is a 128-d vector for each lane. (c) The structure of proposed lane attention module.}
	\label{fig:lane}
\end{figure*}

For each observed trajectory, we use an attention module to predict its future goals, i.e. its target lane in the next 3 seconds. This can effectively guide the network for future predictions. However, the perceived environment is dynamic and there is no fixed number of lanes in different scenarios. To tackle these issues, we apply a dot-product attention module to predict the probability of each lane that the vehicle is heading towards. The architecture of the attention module is shown in Figure~\ref{fig:lane} (c). The ego-vehicle trajectory features will pass through an additional feature embedded layer and then dot-product with the embedded lane features. The output will go through a Softmax layer to obtain the probabilities associated with each lane. This attention module is simple yet flexible to deal with a various number of lanes in the dynamic driving environment. And the probability can be used for downstream tasks including prediction and motion planning.

Suppose that the features for the $n$ possible lanes are $l_1,\cdots, l_n$ respectively. After going through Softmax, the probability for each lane can be written as:
\begin{equation}
    \hat{p}_{i} = \frac{\exp( (W_t^Tf_{{TR}_\text{t}})^T\cdot(W_l^Tf_{l_i}))}{\sum\limits_{j=1}^n\exp((W_t^Tf_{{TR}_\text{t}})^T\cdot(W_l^Tf_{l_j}))}
\end{equation}
where $f_{{TR}_\text{t}}$ is the trajectory features of target vehicle, $\hat{p}_{i}$ is probability for the $i$-th lane, $W_t$ and $W_l$ are the weights of the feature embedding layers for ego-vehicle trajectory and lane, respectively. 

The loss function for the lane intention classification is
\begin{equation}
    \mathcal{L}_{lane} = -\frac{1}{n}\sum\limits_{i=1}^{n}p_i\log(\hat{p}_i),
\end{equation}
where $n$ is the number of lanes.

However, in the real world scenario, only one most possible trajectory can be observed. Training with a single and final ground-truth lane makes the predictions over-confidence. The model tends to predict a single lane, in most cases, it is the current lane, with over 99\% confidence. To alleviate this issue, label smoothing~\cite{ls} is employed. We use $p_s$ for the ground-truth lane and assign other lanes with a probability of equally divided $1-p_s$. This simple method can effectively solve the over-confidence issue and output the reasonable probability for surrounding lanes. In the experiments, we use $p_s=0.8$.

\subsection{Interactions}
Similar to the lane attention module, we model the interactions between agents by using another attention module. Here, the dot-product attention is applied between other agents and the target vehicle. The interaction weight between the target vehicle $v$ and the $j$-th vehicle is  

\begin{equation}
w_{vj} = \frac{\exp[ (W_I^Tf_{{TR}_v})^T\cdot(W_I^Tf_{{TR}_j})]}{\sum\limits_{o}\exp[(W_I^Tf_{{TR}_v})^T\cdot(W_I^Tf_{{TR}_o})]}
\end{equation}
While the final interaction feature $f_{act}$ is

\begin{equation}
f_{act} = \sum\limits_{j\neq v}w_{vj} \cdot f_{TR_j},
\end{equation}

\subsection{Decoder}
The inputs to the decoder include the trajectory feature of the target vehicle $f_{TR_{v}}$, interaction features $f_{act}$ and the feature of the target lane $f_{l_v}$. In the training process, we select the ground-truth lane as input. While during the inference, we can simply feed each lane according to their probability predicted above into the decoder to generate multiple hypothesis. Positions are predicted using a bi-variate Gaussian distribution $\mathcal{N}(x_v^t,y_v^t;\mu_x^t,\mu_y^t,\sigma_x^t,\sigma_y^t,\rho_t)$. The structure of the decoder network is also a standard LSTM network, which output the mean $(\mu_x^t,\mu_y^t)$ and variance $(\sigma_x^t,\sigma_y^t,\rho_t)$ at each time stamp $t$.
The NLL loss is used for training, 
\begin{equation}
    L_{\text{pos}}=-\sum\limits_{t=1}^{t_{\text{h}}} \log P(x_v^t,y_v^t|\mu_x^t,\mu_y^t,\sigma_x^t,\sigma_y^t,\rho_t)
\end{equation}

\subsection{Loss Functions}
There are totally two losses used in our model: lane intention loss and trajectory loss.
The final loss function is:
\begin{equation}
    \mathcal{L} = \mathcal{L}_{lane} + \mathcal{L}_{pos}
\end{equation}

\subsection{Generating Multiple Predictions}
Our lane-attention module can be used for generating multiple predictions and assign each prediction a probability. Specifically, we can simply feed different lane features into the decoder to generate the corresponding prediction. Compared with stochastic multi-modal prediction methods, such as CVAE~\cite{mfp}, our method use the lane feature instead of random noise as the latent variable, which has explicit semantic meaning. Compared with intention-based prediction~\cite{cslstm}, our lane feature can also be viewed as a kind of intention of the agent. Unlike previous methods using a fixed number of intentions such as turning or changing lane, our intention set (i.e. target lanes) is dynamic and is flexible to deal with different driving scenarios. We show qualitative results in the experiment section. 

\section{Experiments}
\begin{table*}[!t]
\centering
\begin{tabular}{|l|l|l|llcccc|l|ll|}
\hline
\multicolumn{3}{|c|}{\multirow{2}{*}{Method}}         & \multicolumn{2}{c}{K=1}                          & \multicolumn{2}{c}{K=3}                             & \multicolumn{2}{c|}{K=6}                             &  & \multicolumn{2}{c|}{NS (K=1)} \\ \cline{4-9} \cline{11-12} 
\multicolumn{3}{|c|}{}                                & \multicolumn{1}{c}{ADE(m)} & \multicolumn{1}{c}{FDE(m)} & \multicolumn{1}{l}{ADE(m)}  & \multicolumn{1}{l}{FDE(m)}  & \multicolumn{1}{l}{ADE(m)}  & \multicolumn{1}{l|}{FDE(m)}  &  & ADE(m)           & FDE(m)           \\ \cline{1-9} \cline{11-12} 
\multicolumn{3}{|l|}{Constant Velocity Kalman Filter} & 2.22                    & 5.09                    & -                        & -                        & -                        & -                         &  & 3.17          & 7.60          \\ \cline{1-9} \cline{11-12} 
\multicolumn{3}{|l|}{NN+map(prune)~\cite{argo}} & 3.38 & 7.62 &  2.11 & 4.36 & 1.68 & 3.19 & & - & - \\ \cline{1-9} \cline{11-12} 
\multicolumn{3}{|l|}{LSTM+map(prior) 1-G,n-C~\cite{argo}} & 2.92 & 6.45 &  2.41 & 4.85 & 2.08 & 4.19 & & - & - \\ \cline{1-9} \cline{11-12} 
\multicolumn{3}{|l|}{MFP~\cite{mfp}} & -                    & -                   & -                        & -                        & 1.39                        & -                         &  & -          & -          \\ \cline{1-9} \cline{11-12} 
\multicolumn{3}{|l|}{~\cite{c7} (Ours implementation)} & 1.60                    & 3.64                   & 1.50                        & 3.21                        & 1.35                        & 2.68                         &  & 1.97          & 4.54          \\ \cline{1-9} \cline{11-12}
\multicolumn{3}{|l|}{LSTM (velocity only)}                          & 1.74                    & 3.98                    & -                        & -                        & -                        & -                         &  & 2.27          & 5.46          \\ \cline{1-9} \cline{11-12} 
\multicolumn{3}{|l|}{LSTM (velocity + position) }                        & 1.61                    & 3.67                    & -                        & -                        & -                        & -                         &  & 2.12          & 5.10          \\ \cline{1-9} \cline{11-12} 
\multicolumn{3}{|l|}{LSTM + Lane}                      & 1.50                    & 3.37                    & \multicolumn{1}{c}{1.30} & \multicolumn{1}{c}{2.81} & \multicolumn{1}{c}{1.08} & \multicolumn{1}{c|}{2.12} &  & 1.90          & 4.37          \\ \cline{1-9} \cline{11-12}
\multicolumn{3}{|l|}{LSTM + Lane + Interact}                    & \textbf{1.46}           & \textbf{3.27}           & \textbf{1.28}            & \textbf{2.78}            & \textbf{1.05}            & \textbf{2.06}             &  & \textbf{1.83} & \textbf{4.23} \\ \hline
\end{tabular}
\caption{Results on the argoverse motion forecasting validation set and a non-straight subset.} 
\label{Tab:hard}
\end{table*}

\begin{table*}[]
\center
\begin{tabular}{|l|llll|llll|llll|}
\hline
\multirow{2}{*}{team} & \multicolumn{4}{c|}{K=1}  & \multicolumn{4}{c|}{K=3}  & \multicolumn{4}{c|}{K=6}  \\ \cline{2-13} 
                      & ADE$\downarrow$  & FDE$\downarrow$  & DAC$\uparrow$  & MR$\downarrow$   & ADE$\downarrow$  & FDE$\downarrow$  & DAC$\uparrow$  & MR$\downarrow$   & ADE$\downarrow$  & FDE$\downarrow$  & DAC$\uparrow$  & MR$\downarrow$   \\ \hline
Jean                  & 1.81 & 4.08 & 0.99 & 0.63 & 1.39 & 2.89 & 0.98 & 0.48 & 0.95 & 1.55 & 0.98 & 0.19 \\ \hline
uulm                  & 1.97 & 4.32 & 0.99 & 0.66 & 1.23 & 2.35 & 0.98 & 0.33 & 0.96 & 1.55 & 0.98 & 0.22 \\ \hline
\textbf{Ours}                  & 1.91 & 4.31 & 0.99 & 0.66 & 1.24 & 2.49 & 0.98 & 0.33 & 0.99 & 1.71 & 0.98 & 0.19 \\ \hline
hale                  & 3.03 & 7.07 & 0.96 & 0.80 & 1.78 & 3.55 & 0.96 & 0.53 & 1.40 & 2.40 & 0.96 & 0.34 \\ \hline
Holmes	& 2.91&	6.54&	1.00&	0.82&	1.78&	3.72&	0.99&	0.59&	1.38&	2.66&	0.99&	0.42 \\ \hline
\end{tabular}
\caption{Results on Argoverse motion forecasting challenge leaderboard (test set). We report the final challenge entries at NeurIPS 2019, while omitting later submissions.}
\label{test}
\end{table*}

\subsection{Dataset and Evaluation Metrics}
Argoverse dataset~\cite{argo} is collected from a fleet of autonomous vehicles in different cities in the USA. In the motion forecasting task, several interesting scenarios, such as vehicles at intersections, taking left or right turns or changing to adjacent lanes, etc. are sampled. In total, 324,557 five-second sequences are collected and used in the forecasting benchmark. Each sequence contains a target vehicle needs to predict, as well as the ego vehicle and other agents within a certain distance. The 324,557 sequences are split into 205,942 train, 39,473 validation, and 78145 test sequences. 
\begin{figure*}[!h]
	\begin{center}
		\includegraphics[width=0.6\linewidth]{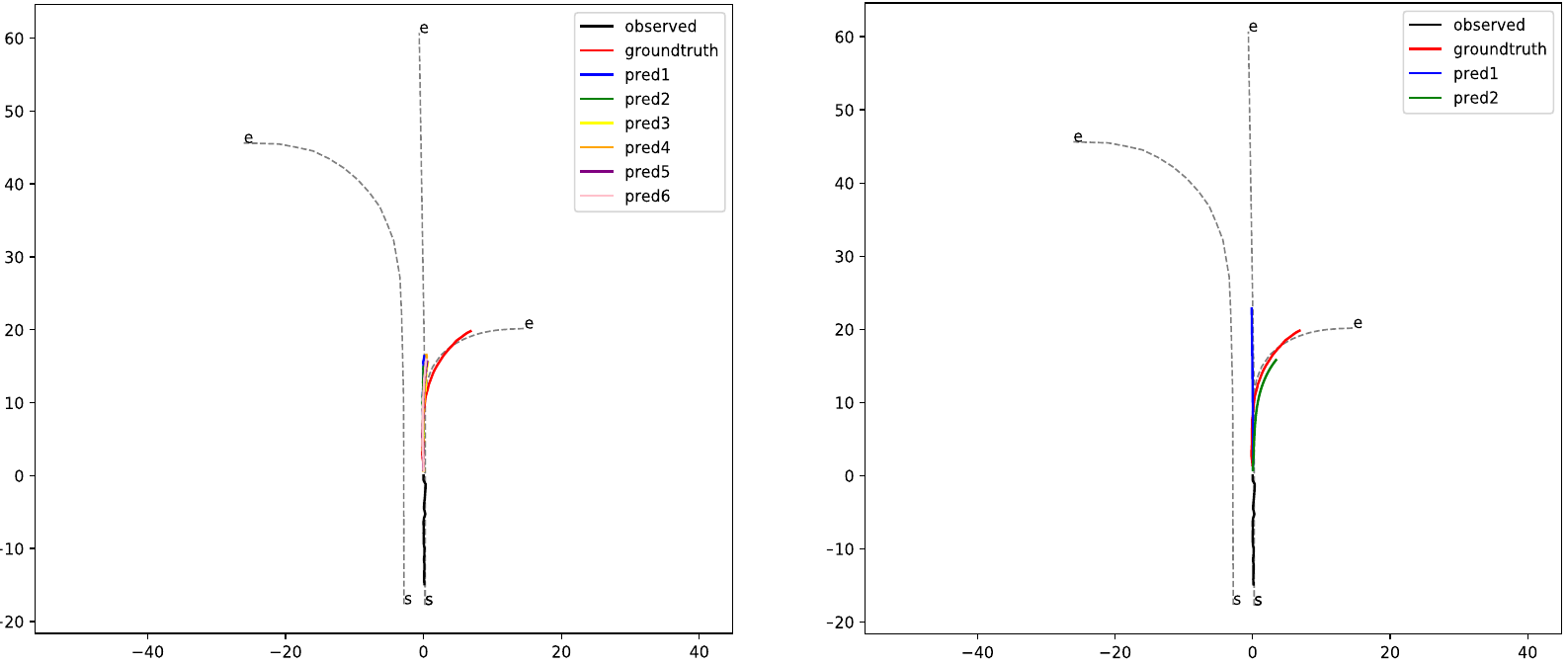}
	\end{center}
	\caption{Left: using rasterized map and directly predict six trajectories~\cite{c7}. Right: our results. We show that using rasterized map may not able to cover all the possibilities}
	\label{fig:compare}
\end{figure*}

 For each trajectory, the past 2s are the observations and we need to predict the future 3s trajectories. We adopt the commonly used metrics: Average Displacement Error (ADE) and Final Displacement Error (FDE), which are defined as
\begin{equation}
ADE=\frac{1}{T}\sum\limits_{t=t_o}^{t_h}\|\hat{Y_t}-Y_t\|,
\end{equation}
and 
\begin{equation}
FDE=\|\hat{Y}_{t_h}-Y_{t_h}\|,
\end{equation}
where $\hat{Y_t}$ are the predicted location at timestamp $t$.

For multi-modal prediction, we use the minADE and minFDE over maximum $K$ predictions, where $K=3, 6$ are used for the Argoverse dataset.

Since in most cases, the cars just follow the current lane and drive straightly, evaluating the whole dataset may be biased. So we select all the non-straight trajectories from the validation set (denoted as ``NS'') based on the provided map and some heuristic rules. This includes lane changing and turning. This subset contains 10,408 sequences, which takes up about 25\% of the whole cases. We also provide a benchmark on these `NS' sequences to vividly present our advantages on non-straight trajectories. 

\subsection{Implementation Details}
We implement the model in Pytorch~\cite{pytorch}. The whole model is trained end-to-end using Adam~\cite{adam} with batchsize of 1024. The initial learning is 1e-2 for about 50 epochs and is decreased to 1e-3 for about 5 epochs. We report the best results of each model on the validation set.

\subsection{Experiment Results}
We report the results on the validation set in Table \ref{Tab:hard}. For the single trajectory evaluation, the prediction with the highest probability will be used. We systematically evaluate each component of our proposed method and compare it with two baselines and one previous state-of-the-art method MFP~\cite{mfp}.

As for trajectory features, encoding both velocity and position is beneficial. Our full model combining lane and interactions can outperform trajectory-only models by a large margin (0.4m for 3s FDE). What's more, the lane attention module allows us to generate multiple hypothesis in a lane-following manner and outperform the stochastic method~\cite{mfp} (0.34m for 3s ADE). 
\begin{figure*}[!t]
	\begin{center}
		\includegraphics[width=0.7\linewidth]{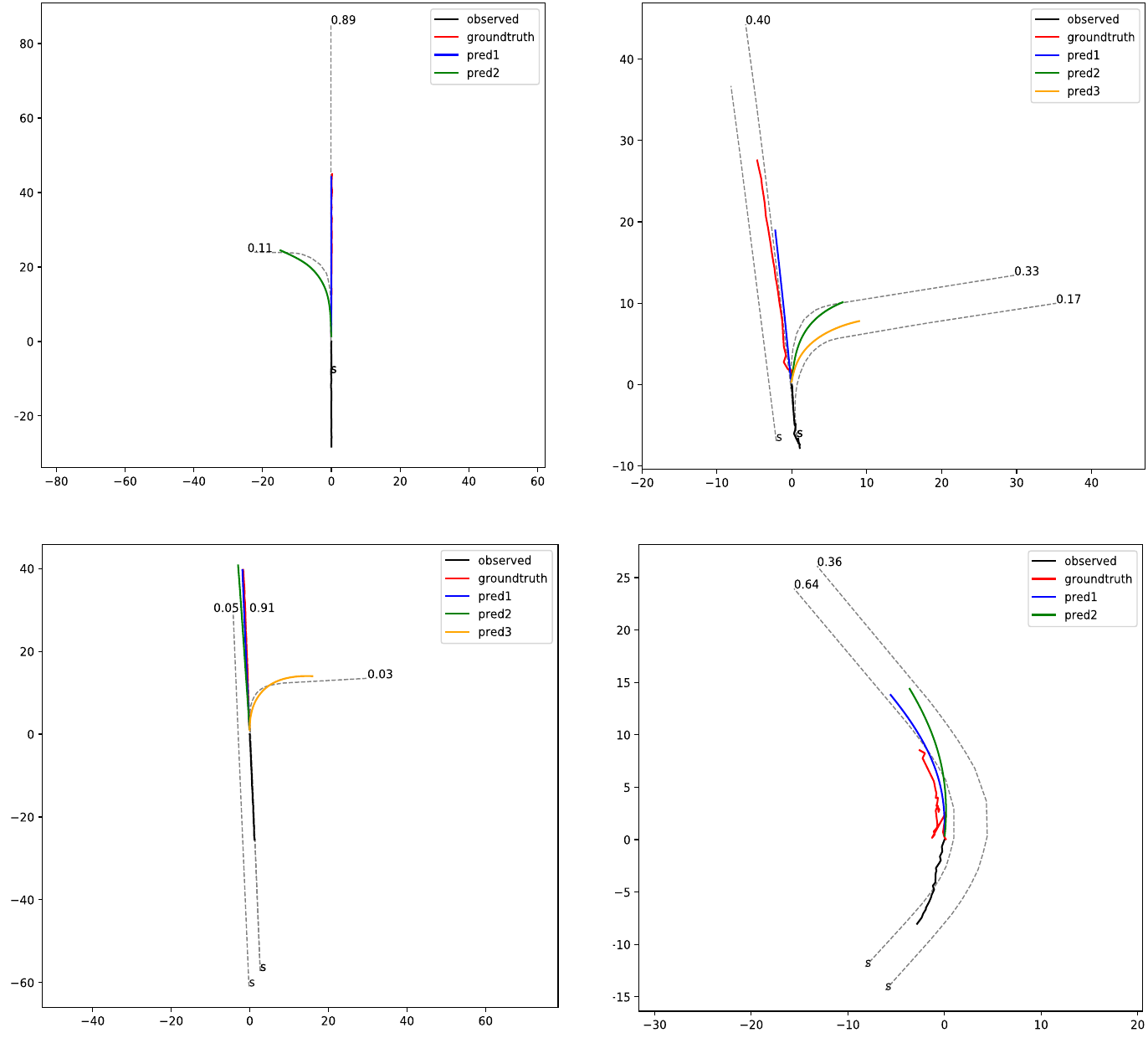}
	\end{center}
	\caption{Examples of our predictions. For each example, we show the probability for each lane and the corresponding predicted trajectory.}
	\label{fig:example}
\end{figure*}

\subsubsection{How Do Lanes Help Prediction}
We test different models trained on the whole training set directly on the ``NS'' subset. The results are shown in Tab.\ref{Tab:hard}. As we can see, the advantage of using lane prior enlarges significantly on the non-straight trajectories compared with performance on all the trajectories. Especially, the final distance error for 3s is nearly 0.9 meters better when using lane information. This shows that the lane information can effectively guide the predictions when the vehicle's state or intention changes. 

\subsubsection{Rasterized Map vs Instance-aware Lane Encoding}
We also implement the method proposed in ~\cite{c7} and report the results in TABLE~\ref{Tab:hard}. Our methods can achieve better results with much a smaller model size. Besides, the holistic map representation may suffer from model collapse, as shown in Fig. \ref{fig:compare}. The six predictions cannot cover the ``turning'', while our instance-aware encoding can explicitly tell the model to cover all the cases. 

\subsubsection{Effectiveness of interactions} In most cases, the trajectory of a vehicle mainly depends on its states and lane topology, adding interaction features only results in marginal improvements. As for the non-straight scenarios such as change lanes or tuning at intersections, observing other vehicles could be necessary. In such cases, the improvements are slightly enlarged by using interaction features. 

\subsubsection{Generating Multiple Predictions}
One advantage of our lane attention module is that it provides a probability for each predicted trajectory. As discussed in Section 1, we can simply feed different lane features into the decoder to generate the corresponding prediction. 

In Figure \ref{fig:example}, we illustrate some examples from our predictions. As we can see, feeding different lane features into the decoder, the model can generate different trajectories. There is a strong correlation between the lane and the prediction. Although during the training, only one ground-truth lane and one ground-truth trajectory are provided, surprisingly, the model can generalize to other lanes. As shown in the example, the model can predict lane-changing behavior. At intersections or forks, the model will output possible trajectories for each lane. 

\subsubsection{Challenge Results}
Here, we present our submission to the Argoverse motion forecasting competition at NeurIPS 2019. For the test submission, we augment the training data by adding all the trajectories from both training and validation set instead of just using the ``agent'' trajectories in the training split. When the number of surrounding lanes is less than six, we sample additional trajectories using the predicted mean $\mu$ and variance $\sigma,\rho$ at each timestamp. 
The results are shown in Table~\ref{test}. For the challenge, two new metrics, DAC and MR are introduced. DAC is Drivable Area Compliance, which is the proportion of predicted trajectories within the drivable area. Miss rate (MR) measures the rate of final displacement greater than 2m. 

As for $K=1,3$, our method ranks on the second place. Note that for $K=6$, we achieve the best result in term of ``MR'' and outperform the later entries by a large margin on all the metrics.

\section{Conclusions}
In this paper, we present a new way for trajectory prediction in the autonomous scenario. Within this architecture, the surrounding lanes provide guidance for future trajectory prediction. Different from other rasterized methods, we instantly sample each surrounding lane coordinate for neural networks to extract the lane information. We show that the lane information can be served as dynamic intentions for generating diverse predictions without suffering from model collapse. 
The quantitative and qualitative results on the current largest public motion forecasting dataset further corroborate the effectiveness of our proposed method.

\bibliographystyle{IEEEtranS}
\bibliography{egbib}

\end{document}